\documentclass[runningheads]{llncs}

\usepackage{eccv}

\usepackage{eccvabbrv}

\usepackage{graphicx}
\usepackage{booktabs}

\usepackage[accsupp]{axessibility}  % Improves PDF readability for those with disabilities.

\usepackage{amsmath}
\usepackage{amsfonts}
\usepackage{bm}
\usepackage{algorithm}
\usepackage{algorithmic}
\usepackage{multirow}
\usepackage{colortbl}

\newcommand{\tabincell}[2]{\begin{tabular}{@{}#1@{}}#2\end{tabular}}
\usepackage{hyperref}

\usepackage{orcidlink}

\begin{document}

\title{
Focusing on What Matters: Saliency-Harnessing Accurate Routing for Diffusion MoE
}

\titlerunning{Saliency-Harnessing Accurate Routing for Diffusion MoE}

% Second Author\inst{2,3}\orcidlink{1111-2222-3333-4444} \and
% Third Author\inst{3}\orcidlink{2222--3333-4444-5555}}

\author{
Haoyou Deng\inst{1,2}\orcidlink{0009-0007-7615-7986} \and
Keyu Yan\inst{2}$^{\dagger}$ \and
Chaojie Mao\inst{2} \and
Xiang Wang\inst{1} \and
Yu Liu\inst{2} \and \\
Changxin Gao\inst{1} \and 
Nong Sang\inst{1}$^{\ddagger}$\orcidlink{0000-0002-9167-1496}
}

\authorrunning{H.~Deng et al.}

% Springer Heidelberg, Tiergartenstr.~17, 69121 Heidelberg, Germany

\institute{
Key Laboratory of Image Processing and Intelligent Control, School of Artificial \\ Intelligence and Automation, Huazhong University of Science and Technology \and
Tongyi Lab, Alibaba Group\\
\email{\{haoyoudeng, wxiang, cgao, nsang\}@hust.edu.cn}\\
\email{yankeyu66@aliyun.com, \{chaojie.mcj, ly103369\}@alibaba-inc.com}
}

\maketitle
{
\renewcommand{\thefootnote}{}
\footnotetext{$^\dagger$Project Leader\quad $^\ddagger$Corresponding Author}
}

\begin{abstract}
 
   Mixture-of-Experts (MoE) architectures have emerged as a powerful paradigm for scaling diffusion models in visual generation. Recent advancements have focused on adaptively allocating computational resources across diverse tokens to improve efficiency and performance.
   However, we identify a routing assignment problem in existing diffusion MoE frameworks: the router fails to accurately allocate more computational resources to salient tokens.
   Our analysis attributes this failure to the router’s reliance on noise-corrupted latent features throughout the denoising process.
   Such stochastic noise obscures the critical structural and textural information, thereby preventing the router from effectively distinguishing salient tokens.
   To address this, we propose \textbf{SharpMoE}, a post-training framework with a saliency-harnessing accurate routing mechanism, which utilizes clean latent features as a noise-free guidance signal for routing.
   By bypassing the noise-distorted inputs, SharpMoE provides the router with clear saliency guidance, enabling the identification of salient tokens even in high-noise stages.
   Furthermore, we introduce a trajectory routing loss to constrain the compute allocation throughout the multi-step denoising trajectory, ensuring precise resource allocation along the generation rollout.
   Extensive experiments demonstrate that SharpMoE serves as a versatile, plug-and-play solution that further enhances the pretrained, converged MoE models, achieving state-of-the-art performance in visual generation.

\end{abstract}

\section{Introduction}
\label{sec:intro}

Diffusion models~\cite{ho2020denoising} have achieved remarkable advancements in visual generation~\cite{rombach2022high, peebles2023scalable, wang2025hbridge, mao2026wanimage}. Recent research has increasingly focused on scaling these models to billions of parameters, with the goal of enhancing image fidelity and generation quality. Diffusion transformers (DiT)~\cite{peebles2023scalable} have emerged as a promising framework, marking a notable architectural shift from U-Net backbones to transformer-based designs and demonstrating exceptional scalability potential. Despite these advancements, however, further scaling DiT models to even larger parameters is hindered by the inherent inefficiency associated with dense parameter activation~\cite{sun2024ecdit}.

To push the boundaries of model scale and capability, the Mixture-of-Experts (MoE) paradigm~\cite{jacobs1991adaptive, shazeer2017outrageously} has emerged as a widely used framework within the large language models (LLMs) community~\cite{zhou2022mixture, liu2024deepseek, li2025minimax, dai2024deepseekmoe, wang2024auxiliary, wang2024remoe}. MoE employs a router that dynamically assigns a sparse subset of parameters (experts) to each input token and aggregates their outputs to produce the final result. This manner enables a significant expansion of model capacity while maintaining computational efficiency.
However, while MoE has achieved substantial success in LLMs, directly applying these established strategies to diffusion models has often resulted in suboptimal performance~\cite{wei2025routing}. This performance gap arises from a fundamental difference in modality: text tokens are discrete and exhibit high semantic density, while visual tokens are spatially correlated and inherently redundant.

To this end, recent research has focused on developing MoE architectures specifically for diffusion-based visual generation.
Early efforts in diffusion MoE~\cite{fei2024tcdit} often relied on the token-choice routing strategy, where each image token is routed to a fixed number of top-ranked experts. However, more recent studies~\cite{sun2024ecdit, shi2025diffmoe, yuan2025expertrace} have identified critical differences in computational requirements across image regions. In diffusion generation, regions containing salient tokens, those rich in critical detail, demand greater computational focus (\eg, more experts) than background or redundant areas. This insight highlights the necessity for a dynamic and saliency-aware routing mechanism capable of adaptively assigning computational resources based on varying saliency levels and textural complexities within an image.

\begin{figure}[t]
    \centering
    \includegraphics[width=1.0\linewidth]{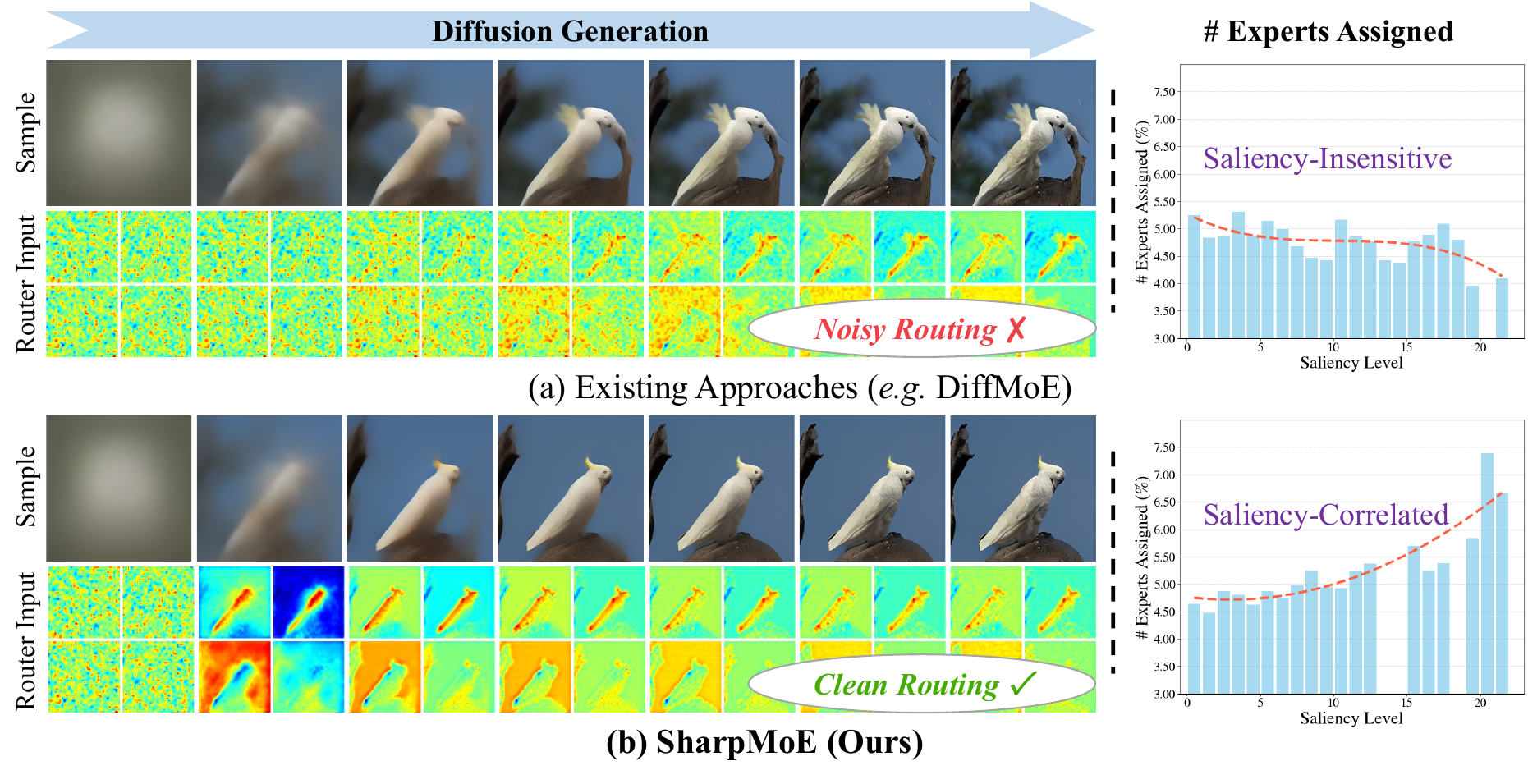}
    \vspace{-0.5em}
    \caption{
    (Left) Visualization of generated samples and per-channel router inputs.
    (Right) Distribution of saliency level and the number of assigned experts.
    (a)  Existing methods struggle to differentiate salient tokens due to the use of noise latents for routing, causing a failure in accurate resource assignment.
    (b) SharpMoE employs clean latents for routing to gain better saliency awareness, thereby achieving a computational allocation that is highly correlated with token saliency.
    }
    % \vspace{-1em}
    \label{fig:motivation}
\end{figure}

Although effective, existing dynamic routing methods, such as DiffMoE~\cite{shi2025diffmoe}, suffer from a routing assignment problem: the router fails to accurately allocate more computational resources to salient tokens. To investigate this issue, we utilize the Laplacian operator to extract the textural information of each token as a saliency representation, and then analyze the distribution of experts assigned to tokens with varying saliency levels.
The findings, depicted in Fig.~\ref{fig:motivation}(a), demonstrate that although existing methods aim to achieve saliency-aware allocation, their actual routing results are largely saliency-insensitive, showing minimal variation in expert allocation across tokens with different saliency levels.
We attribute this limitation to the \textit{noisy routing}, where the router is consistently conditioned on noise-corrupted latents throughout the multi-step denoising process. This pervasive noise, especially pronounced at early high-noise timesteps, masks critical structural and textural details. Ultimately, this corruption impairs the router's ability to effectively distinguish salient regions, which leads to inaccurate allocation of computational resources.

To address the aforementioned issue, we introduce \textbf{SharpMoE}, a post-training framework that incorporates a saliency-harnessing accurate routing mechanism to facilitate \textit{clean routing} for diffusion MoE.
The core insight of SharpMoE is to leverage the clean latents (\ie, the $\hat{\bm{x}}_0$ prediction) from the preceding denoising timestep as the input to the router for the current timestep.
This design offers two distinct advantages:
(1) \textit{Saliency Awareness}: The predicted clean features explicitly capture and highlight the salient regions of the image. As shown in Fig.~\ref{fig:motivation}(b), these features accurately capture the primary object even under the heavy noise present during the early timesteps, delivering robust and well-defined structural guidance to the router.
(2) \textit{Temporal Stability}: The noise-free nature of the latent $\hat{\bm{x}}_0$ ensures a clean and robust input for routing across the entire denoising trajectory. This effectively mitigates the inaccurate resource allocation caused by noise-contaminated inputs in earlier approaches.
Encouragingly, while SharpMoE may seem computationally prohibitive due to the full-trajectory training scheme for obtaining $\hat{\bm{x}}_0$, we demonstrate that it can be efficiently implemented as a post-training algorithm.
Our findings reveal that only a limited number of post-training steps (or gradient updates) are sufficient to deliver significant performance gains, even when applied to fully converged pretrained models, highlighting the efficiency and adaptability of SharpMoE.

Moreover, to further promote a saliency-aware allocation of computational resources, we propose the Trajectory Routing Loss, which drives the assignment of computational effort to align with the underlying saliency distribution. Specifically, leveraging the full-trajectory training paradigm in SharpMoE, we quantify the cumulative computational load for each token by aggregating its expert activations across the denoising sequence.
This approach then regulates the computational budget based on saliency, allowing a precise alignment between resource allocation and the visual significance of different regions. By prioritizing salient tokens, the proposed loss concentrates computational capacity on regions with high structural and textural complexity, thereby significantly improving generative fidelity.
Extensive experiments conducted on multiple pretrained architectures validate the effectiveness of SharpMoE as a plug-and-play post-training enhancement, further emphasizing the crucial role of clean routing in advancing the capabilities of diffusion MoEs.

In summary, our contributions are fourfold:
(1) We identify a critical routing assignment problem in existing diffusion MoEs: the router fails to effectively discriminate salient tokens due to the limitations posed by noisy routing.
(2) To address this, we introduce \textbf{SharpMoE}, a post-training framework that leverages clean latents as noise-free guidance for the router, effectively capturing critical structural and textural cues for accurate saliency identification.
(3) Within SharpMoE, we propose a trajectory routing loss designed to regulate computational resource allocation across the entire generative trajectory, enabling precise saliency-aware routing.
(4) Extensive experiments demonstrate that SharpMoE serves as a plug-and-play enhancement to further boost pretrained diffusion MoE models, achieving state-of-the-art performance in visual generation.
% \vspace{-0.5em}
\section{Related Work}

% \vspace{-0.25em}
\paragraph{Diffusion Models.}
Diffusion models~\cite{ho2020denoising} have demonstrated remarkable success in visual generation~\cite{ma2024sit, wan2025wan, wu2025qwen, deng2026densegrpo, seedream2025seedream, cao2025hunyuanimage}. Early works~\cite{rombach2022high, podell2023sdxl} primarily utilized U-Net~\cite{ronneberger2015unet} backbone optimized via Denoising Diffusion Probabilistic Models (DDPM)~\cite{ho2020denoising, song2020score} objective.
More recently, the field has transitioned toward the Diffusion Transformer (DiT)~\cite{peebles2023scalable} architecture to facilitate model scaling. When combined with the Rectified Flow (RF) training paradigm~\cite{liu2022rectifiedflow}, these DiT-based models~\cite{chen2023pixart, hatamizadeh2024diffit, ma2024sit, wei2024dreamvideo, wang2025hbridge} have demonstrated superior scalability and synthesis quality, setting new benchmarks for high-fidelity generation.

% \vspace{-0.25em}
\paragraph{Mixture of Experts.}
Mixture-of-Experts (MoE) architectures~\cite{shazeer2017outrageously, lepikhin2020gshard} expand model capacity efficiently by leveraging sparse activation, where only a subset of parameters is activated for each token.
While MoE has achieved considerable success in Large Language Models (LLMs), such as DeepSeek-V3~\cite{liu2024deepseek} and MiniMax-01~\cite{li2025minimax}, recent efforts~\cite{balaji2022ediff, xue2023raphael, zhao2024dynamic, fei2024tcdit, sun2024ecdit, shi2025diffmoe, yuan2025expertrace, wei2025routing} have focused on adapting MoE to scale dense diffusion models. However, directly migrating MoE designs from LLMs to diffusion frameworks~\cite{esser2024scaling, fei2024tcdit} often leads to suboptimal results due to modality differences: text tokens are semantically dense, whereas visual tokens are spatially correlated and redundant.
Recent studies have highlighted the heterogeneous complexity of image regions, wherein salient tokens containing critical details demand greater computational resources. This observation has inspired the development of saliency-aware routing mechanisms, such as the expert-choice strategy in EC-DiT~\cite{sun2024ecdit} and the batch-level pooling approach in DiffMoE~\cite{shi2025diffmoe} and Expert Race~\cite{yuan2025expertrace}. However, these mechanisms often struggle to achieve precise expert assignments due to the reliance on noisy latents during the denoising process, which obscure the representation of saliency.
To address this, we present SharpMoE, which leverages predicted clean latents for routing to provide a robust saliency representation for dynamic expert assignment.

\section{Preliminary}
% \vspace{-0.5em}

\paragraph{Diffusion Models.} Diffusion models add Gaussian noise to data and train a neural network to reverse the process.
Let $\bm{x}_0 \sim X_0$ be a sample form the data distribution, and $\bm{x}_1 \sim X_1$ denote a noise sample, the recent advanced Rectified Flow~\cite{liu2022rectifiedflow} framework defines the noised data $\bm{x}_t$ at timestep $t$ as:
\begin{equation}
    \bm{x}_t = t\bm{x}_1 + (1-t)\bm{x}_0.
\end{equation}
Then, a denoising model is trained to directly regress the velocity field $\bm{v}_\theta(\bm{x}_t, t)$ by minimizing the Flow Matching~\cite{lipman2022flow} objective:
\begin{equation}
\label{eq:flow_matching}
    \mathcal{L}(\theta) = \mathbb{E}_{t, \bm{x}_0 \sim X_0, \bm{x}_1 \sim X_1} \left[ \| \bm{v} - \bm{v}_\theta(\bm{x}_t, t) \|^2 \right],
\end{equation}
with the target $\bm{v}=\bm{x}_1-\bm{x}_0$. During generation, the denoising process is formulated as:
\begin{equation}
    \bm{x}_{t+dt} = \bm{x}_{t} + dt \cdot \bm{v}_\theta(\bm{x}_{t}, t),
\end{equation}
where $dt$ is the timestep gap. Therefore, the $\hat{\bm{x}}_0$ prediction at timestep $t$ is:
\begin{equation}
    \hat{\bm{x}}_0 = \bm{x}_{t} -t \bm{v}_\theta(\bm{x}_{t}, t).
\end{equation}

\paragraph{Mixture of Experts.}
The Mixture-of-Experts (MoE) adaptively activates a sparse subset of ``experts'' for each token. In general, a standard MoE layer consists of a router $\mathcal{R}$ and $N_E$ experts $\{E_i \}_{i=1}^{N_E}$, each implemented as a Feed-Forward Network (FFN). Given an input $\bm{x} \in \mathbb{R}^{B\times S \times D}$, where $B$, $S$, and $D$ denote the batch size, token length, and hidden dimension, respectively, the router $\mathcal{R}$ predicts a set of token–expert affinity scores $\bm{S} \in \mathbb{R}^{B\times S \times N_E}$:
\begin{equation} \label{eq:router}
    \bm{S}=\mathcal{R}(\bm{x}).
\end{equation}
Subsequently, the router selects the experts with the top-k highest scores for computation, and the MoE output is the weighted sum of these experts’ outputs:
\begin{align}
    \bm{G} = \begin{cases}\bm{S}, & \text { if } \bm{S} \in \operatorname{TopK}(\bm{S}, K) \\
    0, & \text { Otherwise }\end{cases}, 
    \quad
    \operatorname{MoE}(\bm{x}) = \sum_{i=1}^{N_E} \bm{G}_i * E_i(\bm{x}),
    \label{eq:moe_output}
\end{align}
where $\bm{G} \in \mathbb{R}^{B \times S \times N_E}$ is the final gating tensor, and $\operatorname{TopK}(\cdot, K)$ is the operation that selects a subset with the $K$ largest value.
Notably, the current input $\bm{x}$ in Eq.~\ref{eq:router} is corrupted by the remaining noise during diffusion generation, raising a noisy routing issue where the router fails to accurately differentiate salient tokens, ultimately resulting in incorrect routing assignments.
\begin{figure}[t]
    \centering
    \includegraphics[width=0.85\linewidth]{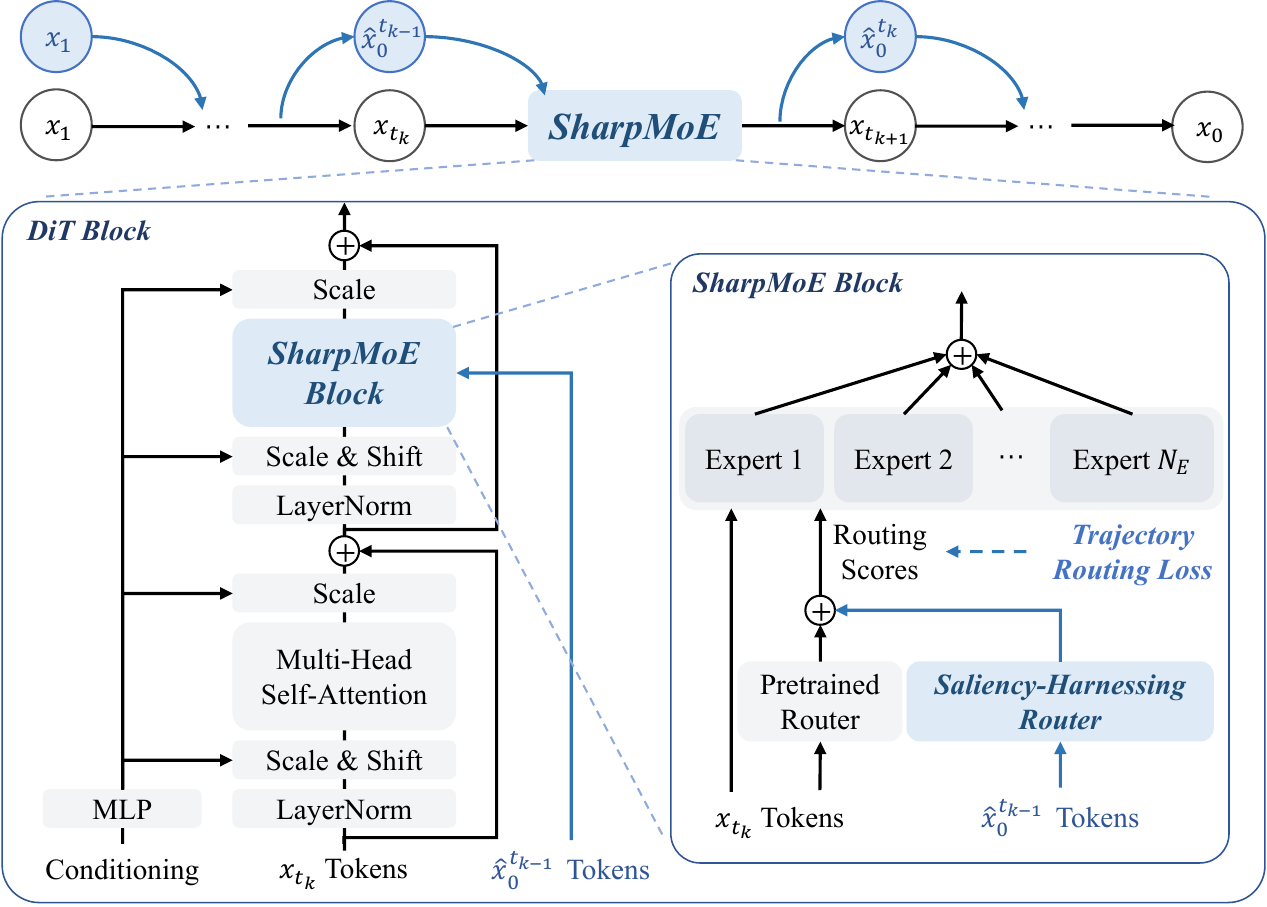}
    \caption{
    Overview of SharpMoE architecture. SharpMoE leverages a full-trajectory training scheme where predicted clean latents $\hat{\bm{x}}_0$ provide saliency guidance for routing. Within each SharpMoE Block, a Saliency-Harnessing Router (taking $\hat{\bm{x}}_0^{t_{k-1}}$) complements the standard Pretrained Router (taking $\bm{x}_{t_k}$) for precise expert assignment. The Trajectory Routing Loss is then imposed on the routing scores to align the cumulative compute allocation with the image's saliency distribution across all denoising steps.
    }
    \label{fig:framework}
    
\end{figure}

\section{SharpMoE}

\subsection{Overview}

We build SharpMoE upon the DiT architecture~\cite{peebles2023scalable}, replacing standard feed-forward networks (FFNs) with SharpMoE blocks to facilitate scalable modeling with high efficiency. The core idea of SharpMoE is to transition from \textit{noisy routing} to \textit{clean routing} by harnessing saliency, ensuring that computational resources dynamically focus on salient regions during the denoising process.

As depicted in Fig.~\ref{fig:framework}, SharpMoE introduces three key components to achieve this goal:
(1) At a timestep $t_k$, we depart from the conventional reliance on noisy latents $\bm{x}_{t_k}$ for routing decisions. Instead, we propose a Saliency-Harnessing Router mechanism (Sec.~\ref{sec:router}) that incorporates the clean prediction $\hat{\bm{x}}_0^{t_{k-1}}$ from the preceding timestep $t_{k-1}$, providing robust saliency information.
(2) The introduction of $\hat{\bm{x}}_0^{t_{k-1}}$ creates a recursive dependency across timesteps. To facilitate the availability of these clean latents during training, we develop a Recursive Full-Trajectory Training scheme (Sec.~\ref{sec:full_trajectory_training}). Unlike standard single-step denoising, this training strategy enables the model to learn and utilize recursive dependencies across the entire generative trajectory.
(3) To regulate cumulative expert allocation, we introduce the Trajectory Routing Loss (Sec.~\ref{sec:routing_loss}). By exploiting the global perspective offered by this training paradigm, this loss aligns the total computational resources with the image's saliency distribution across all timesteps.
In the following subsections, we detail each of these components within the SharpMoE framework.

\subsection{Saliency-Harnessing Accurate Routing}
\label{sec:router}

In diffusion-based generation, visual tokens exhibit non-uniform informational density. Salient tokens, which represent critical object structures and complex textures, necessitate a higher computational budget (\ie, more experts) to ensure generative fidelity. Conversely, redundant background tokens can be processed with fewer experts. While current diffusion MoE frameworks~\cite{sun2024ecdit, shi2025diffmoe} attempt saliency-aware resource allocation, they primarily rely on the current noisy latent $\bm{x}_{t_k}$ for routing decisions. This reliance gives rise to the \textit{noisy routing} problem: the underlying semantic saliency is heavily obscured by remaining noise, particularly during early timesteps (high noise levels). As a result, routing decisions become erratic and suboptimal, with expert assignments often failing to effectively prioritize computational resources for salient regions.

To tackle the issue of noisy routing, we propose to harness the predicted clean latent $\hat{\bm{x}}_0^{t_{k-1}}$ as a saliency representation to provide a stable and noise-free guidance signal for routing.
This design choice is grounded in two key observations:
First, the latent space of the variational autoencoder (VAE) is trained to encode visual information into a semantically dense representation, inherently emphasizing structurally significant regions. As evidenced by Fig.~\ref{fig:motivation}, these latent features naturally encapsulate high-level semantic information, such as object regions and textural complexity, both of which are direct indicators of visual saliency.
Second, within the diffusion framework, $\hat{\bm{x}}_0^{t_{k-1}}$ represents the model's denoised projection onto the clean image manifold at timestep $t_{k-1}$. As visualized in Fig.~\ref{fig:motivation}(b), unlike the noise-perturbed $\bm{x}_{t_k}$ progressively recovers the global structural layout and local details in a denoising manner, $\hat{\bm{x}}_0^{t_{k-1}}$ provides a stable, noise-free estimation of the image's semantic skeleton, even at early stages where local textures are yet to emerge. By routing based on $\hat{\bm{x}}_0^{t_{k-1}}$, SharpMoE effectively mitigates the adverse effects of residual noise, enabling the model to accurately anticipate and prioritize salient regions with high precision.

As shown in Fig.~\ref{fig:framework}, the SharpMoE block employs a dual-router mechanism to integrate saliency-aware information into routing assignment, comprising a pretrained router $\mathcal{R}_{pre}$ and a saliency-harnessing router $\mathcal{R}_{sal}$.
This architecture is designed as a plug-and-play post-training enhancement for established pretrained diffusion MoE models, enabling a seamless integration of saliency information into existing diffusion MoE blocks.
In this setup, the original router from the pretrained backbone is retained as $\mathcal{R}_{pre}$, which assesses the current generation state $\bm{x}_{t_k}$ and thereby captures the transient requirements of the denoising process at timestep $t_k$.
In parallel, we introduce our saliency-harnessing router $\mathcal{R}_{sal}$. By processing the predicted clean tokens $\hat{\bm{x}}_0^{t_{k-1}}$ from the preceding step, $\mathcal{R}_{sal}$ embeds saliency-aware guidance into the routing process, aligning expert allocation with the underlying semantic structure of the image. The final routing scores $\bm{S}$ are derived by fusing the outputs of both routers:
\begin{equation}
    \bm{S}=\mathcal{R}_{pre}(\bm{x}_{t_k})+\mathcal{R}_{sal}(\hat{\bm{x}}_0^{t_{k-1}}).
\end{equation}
To ensure the smooth integration of saliency-aware guidance, we initialize the weights of $\mathcal{R}_{sal}$ to zero, allowing the MoE model to progressively incorporate saliency-aware guidance without disrupting its established denoising capabilities in the post-training stage.

\begin{algorithm}[t]
    \caption{Recursive Full-Trajectory Training for SharpMoE}
    \label{alg:sharpmoe_training}
    \centering
    \begin{algorithmic}[1]
    \STATE {\bf Input:} SharpMoE model ${v}_\theta$ with saliency-harnessing router $\mathcal{R}_{sal}$, Pre-trained model weights $\mathcal{W}_{pre}$, Rollout steps $T$, Loss hyperparameter $\lambda_{routing}$.
    \STATE {\bf Initialize:} 
    \STATE $\quad$ $\theta_{\mathcal{R}_{sal}} \gets 0$ \quad \COMMENT{Initialize saliency-harnessing router with zero}
    \STATE $\quad$ $\theta_{others} \gets \mathcal{W}_{pre}$ \quad \COMMENT{Initialize other parts with pre-trained weights}
    \WHILE{not converged}
        \STATE Sample clean data $\bm{x}_0 \sim X_0$ and noise $\bm{x}_1 \sim X_1$
        % \STATE \COMMENT{Sample $N$ random descending timesteps for the current iteration}
        \STATE Sample $\{t_k\}_{k=1}^T \subset \mathcal{U}(0, 1)$ and sort such that $t_1 > t_2 > \dots > t_T$
        \STATE Set $t_1 = 0.999$
        % \STATE $\hat{\bm{x}}_0^{t_0} = \bm{x}_{t_1}$ \quad \COMMENT{Cold-start: use noise latent as saliency proxy}
        \STATE $\mathcal{L}_{total} = 0$
        \FOR{$k = 1$ {\bf to} $T$}
            \STATE $\bm{x}_{t_k} = t_k \bm{x}_1 + (1 - t_k) \bm{x}_0$ \quad \COMMENT{Initial noisy latent}
        
            \IF{$k=1$}
                \STATE $\hat{\bm{x}}_0^{t_0} = \bm{x}_{t_1}$ \quad \COMMENT{First step: use noise latent as saliency proxy}
            \ENDIF
            
            \STATE $\bm{v}_k = v_\theta(\bm{x}_{t_k}, t_k, \text{sg}(\hat{\bm{x}}_0^{t_{k-1}}))$ \quad \COMMENT{Recursive rollout, $\text{sg}(\cdot)$: stop-gradient}
            \STATE $\mathcal{L}_{fm} = \| (\bm{x}_1 - \bm{x}_0) - \bm{v}_k \|^2$ \quad \COMMENT{Flow Matching objective}
            \STATE $\mathcal{L}_{total} \leftarrow \mathcal{L}_{total} + \mathcal{L}_{fm}$
            \STATE $\hat{\bm{x}}_0^{t_k} = \bm{x}_{t_k} - t_k \bm{v}_k$ \quad \COMMENT{Derive clean prediction for next step}
            \STATE Collect routing scores $\bm{S}_k$
        \ENDFOR
        \STATE Calculate $\mathcal{L}_{routing}$ with $\{\bm{S}_k\}$ \quad \COMMENT{Calculate trajectory routing loss}
        \STATE $\mathcal{L}_{total} \leftarrow \frac{1}{T}\mathcal{L}_{total} + \lambda_{routing}\mathcal{L}_{routing}$
        \STATE Update parameters $\theta$ by minimizing $\mathcal{L}_{total}$
    \ENDWHILE
    \RETURN ${v}_\theta$
    \end{algorithmic}
\end{algorithm}

\subsection{Recursive Full-Trajectory Training}
\label{sec:full_trajectory_training}

Standard diffusion training typically follows a single-step denoising paradigm. Yet, this approach is inherently incompatible with SharpMoE, as the saliency-harnessing router $\mathcal{R}_{sal}$ requires the predicted clean latent $\hat{\bm{x}}_0^{t_{k-1}}$ from the preceding denoising step to provide saliency guidance. Since this preceding state is never computed in a standard single-step setting, $\mathcal{R}_{sal}$ becomes impractical to optimize. This intrinsic recursive dependency necessitates a transition from single-step training to a Recursive Full-Trajectory Training scheme.

As depicted at the top of Fig.~\ref{fig:framework}, each training iteration simulates a short-range generation rollout by sampling $T$ consecutive timesteps $\{t_k\}_{k=1}^{T} \subset[0,1]$, where $t_1>t_2>\dots>t_T$.
At each timestep $t_k$, SharpMoE processes the current noisy latent $\bm{x}_{t_k}$ alongside the predicted clean latent $\hat{\bm{x}}_0^{t_{k-1}}$ from the preceding step to estimate the velocity field $\bm{v}_k$. Then we derive the subsequent latent state $\bm{x}_{t_{k+1}}$ and the clean prediction $\hat{\bm{x}}_0^{t_{k}}$:
\begin{equation}
    \bm{x}_{t_{k+1}} = \bm{x}_{t_{k}} + (t_{k+1}-t_k)\cdot \bm{v}_k, \quad \hat{\bm{x}}_0^{t_{k}}=\bm{x}_{t_{k}}-t_k\cdot \bm{v}_k.
\end{equation}
These outputs are subsequently propagated to the next sampling step, enabling a recursive full-trajectory generative rollout throughout the training process.

A practical challenge arises during inference at the initial timestep $t=1$, where no prior $\hat{\bm{x}}_0$ is available for the saliency-harnessing router $\mathcal{R}_{sal}$. To resolve this, we use the noise latent $\bm{x}_{1}$ as a proxy for the saliency guidance signal. This is motivated by the intuition that during the earliest stage of generation, the object structure and its corresponding saliency are yet to be determined, making the noisy latent a reasonable starting point.
To ensure consistency between the training and inference stages, we ideally set the first timestep $t_1$ of the rollout to $1$. In practice, we set $t_1=0.999$ as a near-equivalent approximation, as initializing with pure Gaussian noise leads to an unconstrained generation target, which prevents the objective in Eq.~\ref{eq:flow_matching} from providing meaningful training signals. The complete recursive full-trajectory training scheme is detailed in Algorithm~\ref{alg:sharpmoe_training}.

\subsection{Trajectory Routing Loss}
\label{sec:routing_loss}

% By aggregating these allocations over the entire trajectory, this approach allows for the quantification of the cumulative computational focus assigned to individual tokens, thereby enabling a trajectory-wise optimization of expert resource utilization.
% By aggregating assignments across the entire rollout, we obtain a more faithful measure of the resources allocated to each token, ensuring precise alignment with visual importance.

To enhance saliency-aware allocation of computational resources, we introduce the Trajectory Routing Loss ($\mathcal{L}_{routing}$), which explicitly aligns the cumulative expert allocation with the saliency distribution.
Building upon the full-trajectory training scheme described above, we observe a notable advantage in its ability to provide a holistic perspective on expert allocation throughout the entire generation process.
Unlike conventional methods limited to single timesteps, this global view better reflects the sequential nature of diffusion, where single-step snapshots may offer a biased estimate of the total computational load for each token.
By considering the entire trajectory, we obtain a more faithful measure of resource distribution.
Specifically, for a $T$-step rollout sequence, we compute the total trajectory-level assignment scores $\mathcal{A}_i$ for the $i$-th token by aggregating the routing scores of its assigned experts over all steps, as follows:
\begin{equation}
    \mathcal{A}_i = \sum_{k=1}^{T} \sum_{l=1}^{L} \sum_{e=1}^{N_E} \mathcal{I}(k, l, e, i) \bm{S}_{k,l}(e,i),
\end{equation}
where $\mathcal{I}(k, l, e, i)$ is an indicator function specifying whether the $i$-th image token is assigned with the $e$-th expert in the $l$-th layer at the $k$-th denoising step, and $\bm{S}_{k,l}(e,i)$ denotes the corresponding token–expert affinity scores for routing. $L$ is the number of MoE layers, each containing $N_E$ experts.

Meanwhile, we adopt the Laplacian operator to estimate the saliency level of an image. In generative modeling, high-frequency components, comprising intricate textures, sharp edges, and foreground boundaries, are inherently coupled with visual saliency. These regions represent critical structural details that necessitate higher numerical precision and more assigned experts. By calculating the second-order derivatives of the clean image, the Laplacian response effectively isolates areas of high structural density, providing a robust representation for saliency levels. Hence, the target saliency map $\mathcal{M}$ is obtained by passing the clean image $\mathbf{X}_0$ through the Laplacian operator $\nabla^2$:
\begin{equation}
    % \mathcal{M}_i = \frac{1}{|P_i|} \sum_{p \in P_i} \nabla^2 \bm{x}_0(p).
    \mathcal{M} = \text{AvgPool} \left( \nabla^2 \mathbf{X}_0  \right),
\end{equation}
where the $i$-th element $\mathcal{M}_i$ represents the saliency level for the $i$-th token.
To ensure that computational resources are allocated in proportion to the saliency level of each token, we minimize the Kullback-Leibler (KL) divergence between the normalized allocation and saliency:
\begin{equation}
\mathcal{L}_{routing} = D_{KL} \left( \text{softmax}(\mathcal{A} ) \parallel \text{softmax}(\mathcal{M}) \right).
\end{equation}
This loss function encourages SharpMoE to prioritize tokens with high saliency throughout the entire generative process. As a result, computational redundancy in less significant background regions is effectively reduced, while the fidelity of crucial foreground details is significantly enhanced.
The overall training objective is formulated as a weighted combination of the Flow Matching loss $\mathcal{L}_{fm}$ (Eq.~\ref{eq:flow_matching}) and the trajectory routing loss:
\begin{equation} \label{eq:total_loss}
\mathcal{L}_{total} = \mathcal{L}_{fm} + \lambda _{routing}\mathcal{L}_{routing}.
\end{equation}
Here, $\lambda_{routing}$ is a balancing hyperparameter that controls the strength of the routing constraint, which is set to $0.001$ in our experiments.

\section{Experiment}

\begin{table}[t]
  \centering
    \caption{
    Quantitative comparison on ImageNet (256$\times$256). We report FID and IS for models undergoing 100K post-training steps following an initial pre-training phase of 500K steps. All evaluations are conducted under RF with CFG scales of 1.0 and 1.5.
    }
  \footnotesize
  \setlength{\tabcolsep}{1.5mm}{
    \begin{tabular}{lcccccc}
    \toprule
    \multirow{2}{*}{Model} 
    & \multirow{2}{*}{\# \tabincell{c}{Activated\\Params.}} 
    & \multirow{2}{*}{\# \tabincell{c}{Total\\Params.}}
    & \multicolumn{2}{c}{cfg=1.0}
    & \multicolumn{2}{c}{cfg=1.5}
    \\
    \cmidrule(lr){4-5} \cmidrule(lr){6-7}
    & & & FID50K $\downarrow$ & IS $\uparrow$ & FID50K $\downarrow$ & IS $\uparrow$ \\
    \midrule
    Dense-DiT-S & 33M & 33M & 55.12 & 26.17 & 27.05 & 58.48 \\
    Dense-DiT-B & 130M & 130M & 31.87 & 47.02 & 10.07 & 122.15 \\
    Dense-DiT-L & 458M & 458M & 19.02 & 71.61 & 4.74 & 183.54 \\
    \midrule
    TC-DiT-S & 33M & 83M & 53.37 & 27.14 & 25.82 & 60.74 \\
    \rowcolor{cyan!10} +SharpMoE & 35M & 85M & \textbf{46.87} & \textbf{31.10} & \textbf{20.55} & \textbf{73.94} \\
    TC-DiT-B & 130M & 329M & 36.70 & 41.04 & 12.82 & 104.37 \\
    \rowcolor{cyan!10} +SharpMoE & 137M & 336M & \textbf{31.61} & \textbf{48.91} & \textbf{9.28} & \textbf{130.52} \\
    TC-DiT-L & 458M & 1.16B & 20.53 & 69.01 & 5.07 & 174.98 \\
    \rowcolor{cyan!10} +SharpMoE & 484M & 1.18B & \textbf{16.63} & \textbf{81.29} & \textbf{3.72} & \textbf{206.93} \\
    \midrule
    EC-DiT-S & 33M & 83M & 50.13 & 28.54 & 23.65 & 64.63\\
    \rowcolor{cyan!10} +SharpMoE & 35M & 85M & \textbf{48.78} & \textbf{30.15} & \textbf{21.81} & \textbf{70.56} \\
    EC-DiT-B & 130M & 329M & 30.35 & 48.90 & 9.28 & 126.60 \\
    \rowcolor{cyan!10} +SharpMoE & 137M & 336M & \textbf{28.10} & \textbf{54.14} & \textbf{7.67} & \textbf{144.50} \\
    EC-DiT-L & 458M & 1.16B & 16.61 & 78.20 & 4.09 & 195.12 \\
    \rowcolor{cyan!10} +SharpMoE  & 484M & 1.18B & \textbf{14.82} & \textbf{88.02} & \textbf{3.27} & \textbf{221.36} \\
    \midrule
    DiffMoE-S & 33M & 140M & 46.55 & 31.96 & 20.59 & 73.46 \\
    \rowcolor{cyan!10} +SharpMoE & 34M & 141M & \textbf{45.52} & \textbf{33.73} & \textbf{19.04} & \textbf{80.07} \\
    DiffMoE-B & 130M & 559M & 27.50 & 54.45 & 8.03 & 138.49 \\
    \rowcolor{cyan!10} +SharpMoE & 133M & 562M & \textbf{25.71} & \textbf{59.60} & \textbf{6.66} & \textbf{155.64} \\
    DiffMoE-L & 458M & 1.98B & 15.13 & 83.62 & 3.86 & 203.00 \\
    \rowcolor{cyan!10} +SharpMoE & 470M & 1.99B & \textbf{13.93} & \textbf{93.66} & \textbf{3.10} & \textbf{228.88} \\
    \bottomrule
    \end{tabular}
    }
  \label{tab:main_result}
\end{table}

\subsection{Experiment Setup}

\paragraph{Baseline and Model Architecture.}
We compare against Dense-DiT~\cite{peebles2023scalable}, and diffusion MoE approaches, including TC-DiT~\cite{fei2024tcdit}, EC-DiT~\cite{sun2024ecdit}, and DiffMoE~\cite{shi2025diffmoe}. All the methods are evaluated across three standardized model scales (S, B, and L). Building on the pretrained states of these methods, SharpMoE serves as a post-training method by incorporating the proposed saliency-harnessing router into the MoE layers of these diffusion MoE models. Each saliency-harnessing router is designed as a two-layer MLP with SiLU activations. Further architectural details are presented in Sec.~A of the supplemental material.

\paragraph{Implementation Detail.}
Following~\cite{shi2025diffmoe}, we perform class-conditional image generation at a resolution of $256\times256$ using the ImageNet~\cite{deng2009imagenet} dataset, which contains 1,281,167 training images across 1,000 classes.
All models are trained within the Rectified Flow (RF) paradigm \cite{liu2022rectifiedflow}, optimized using AdamW with a learning rate of $1\times10^{-4}$ and a batch size of 256.
Besides, we adopt an Exponential Moving Average (EMA) of the model parameters with a decay rate of 0.9999, and all quantitative results reported in this study are computed using the EMA weights.
For SharpMoE, we utilize the full-trajectory training scheme with $T=10$ sampling steps and optimize all network parameters using the loss $\mathcal{L}_{total}$ introduced in Eq.~\ref{eq:total_loss}. Notably, as TC-DiT distributes computational resources uniformly to each token, $\mathcal{L}_{routing}$ becomes inapplicable, leading to training being performed exclusively with $\mathcal{L}_{fm}$.

\paragraph{Evaluation Metric.}
We evaluate the image generation quality of all methods using the Fréchet Inception Distance (FID)~\cite{heusel2017gans, dhariwal2021diffusion} metric, computed over 50,000 generated samples with 250 sampling steps via Flow Matching Euler. Additionally, we report the Inception Score (IS)~\cite{salimans2016improved} to evaluate the diversity of the generated images. A lower FID and a higher IS indicate better performance.

\begin{figure}[t]
    \centering
    \includegraphics[width=1.0\linewidth]{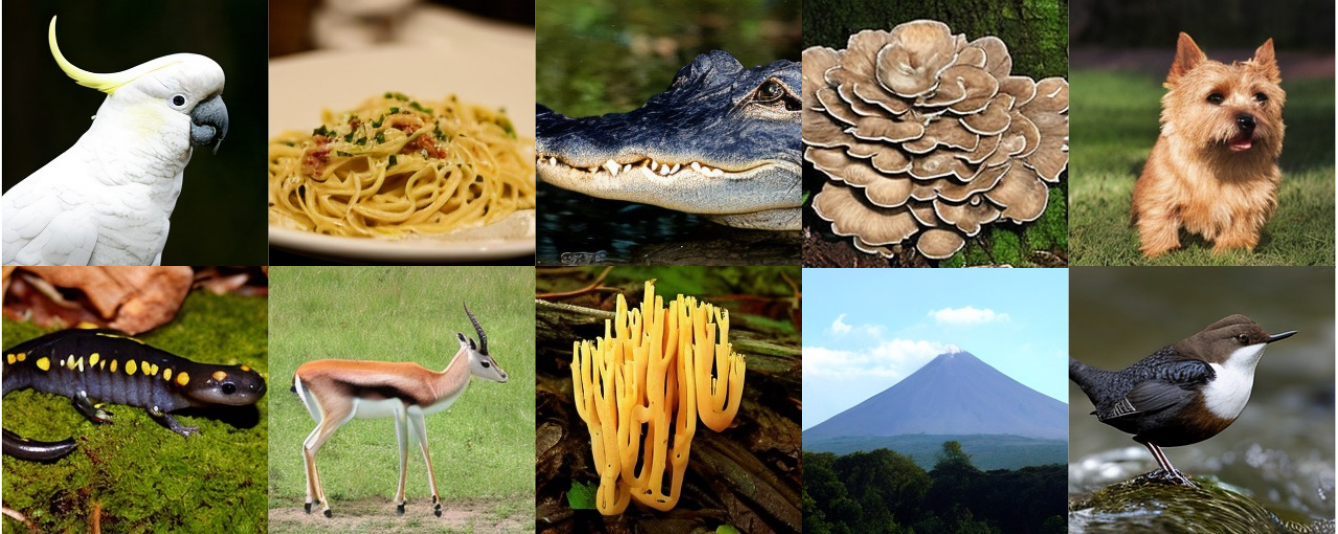}
    \caption{Samples generated by SharpMoE after 100K post-training steps, based on 500K-pretrained DiffMoE-L, with cfg=4.0.}
    \label{fig:main_result_vis}
\end{figure}

\begin{table}[t]
    \centering
    \caption{Ablation study of each component on ImageNet (256$\times$256), evaluated with CFG scales of 1.0 and 1.5.}
    \footnotesize
    \setlength{\tabcolsep}{3mm}{
        \begin{tabular}{lcccc}
            \toprule
            \multirow{2}{*}{Model} & \multicolumn{2}{c}{cfg=1.0} & \multicolumn{2}{c}{cfg=1.5} \\
            \cmidrule(lr){2-3} \cmidrule(lr){4-5}
            & FID50K $\downarrow$ & IS $\uparrow$ & FID50K $\downarrow$ & IS $\uparrow$ \\
            \midrule
            DiffMoE-B & 27.50 & 54.45 & 8.03 & 138.49 \\
            + Saliency-Harnessing Routing & 25.93 & 58.33 & 6.95 & 152.65 \\
            + Trajectory Routing Loss & \textbf{25.71} & \textbf{59.60} & \textbf{6.66} & \textbf{155.64} \\
            \bottomrule
        \end{tabular}
    }
    \label{tab:ablation_component}
\end{table}

\subsection{Main Result}

We evaluate the performance of SharpMoE against other baselines after 100K post-training steps, with all models initialized from pre-trained checkpoints obtained after 500K training steps. As summarized in Tab.~\ref{tab:main_result}, SharpMoE consistently outperforms all competing Diffusion MoE methods, including TC-DiT, EC-DiT, and DiffMoE, across all evaluation metrics, model scales (S, B, and L), and classifier-free guidance (CFG)~\cite{ho2022classifier} scales. This consistent superiority demonstrates the broad effectiveness of our saliency-harnessing routing mechanism across diverse MoE-based diffusion architectures. Importantly, these substantial performance improvements are achieved in just 100K post-training iterations, highlighting SharpMoE's efficiency as a versatile, plug-and-play framework to enhance pretrained models, even when they have already converged.

Among all configurations, SharpMoE achieves its strongest performance when applied to the DiffMoE-L backbone, attaining an FID score of 3.10 and an IS of 228.88 using $\text{cfg}=1.5$. Additionally, even with TC-DiT, which employs a uniform computational resource allocation strategy across all tokens, SharpMoE demonstrates its ability to optimize expert-token assignments by leveraging saliency cues, further improving performance. These quantitative improvements are further validated by the qualitative results presented in Fig.~\ref{fig:main_result_vis}, where SharpMoE-enhanced models demonstrate exceptional structural fidelity and richer textural details, thereby affirming the effectiveness of the proposed saliency-harnessing routing mechanism.

\subsection{Analysis}

\paragraph{Effect of Each Component.}
We conduct an ablation study on ImageNet ($256\times256$) to evaluate the contribution of each component in SharpMoE. The results, summarized in Tab.~\ref{tab:ablation_component}, reveal that integrating the saliency-harnessing routing mechanism leads to a significant performance improvement, reducing the FID from 8.03 to 6.95 at $\text{cfg}=1.5$. This highlights the critical role of clean latent guidance in enhancing saliency awareness compared to conventional noisy routing methods. Additionally, incorporating the trajectory routing loss further improves the FID to 6.66, underscoring the effectiveness of globally aligning cumulative expert allocation with the saliency distribution of the image. This alignment enables sharper focus of computational resources on the most critical regions, improving overall performance. Together, these complementary components achieve the highest generative fidelity across all CFG scales.

\paragraph{Effect of Pretrained Stage.}

\begin{figure}[t]
    \centering
    \includegraphics[width=0.95\linewidth]{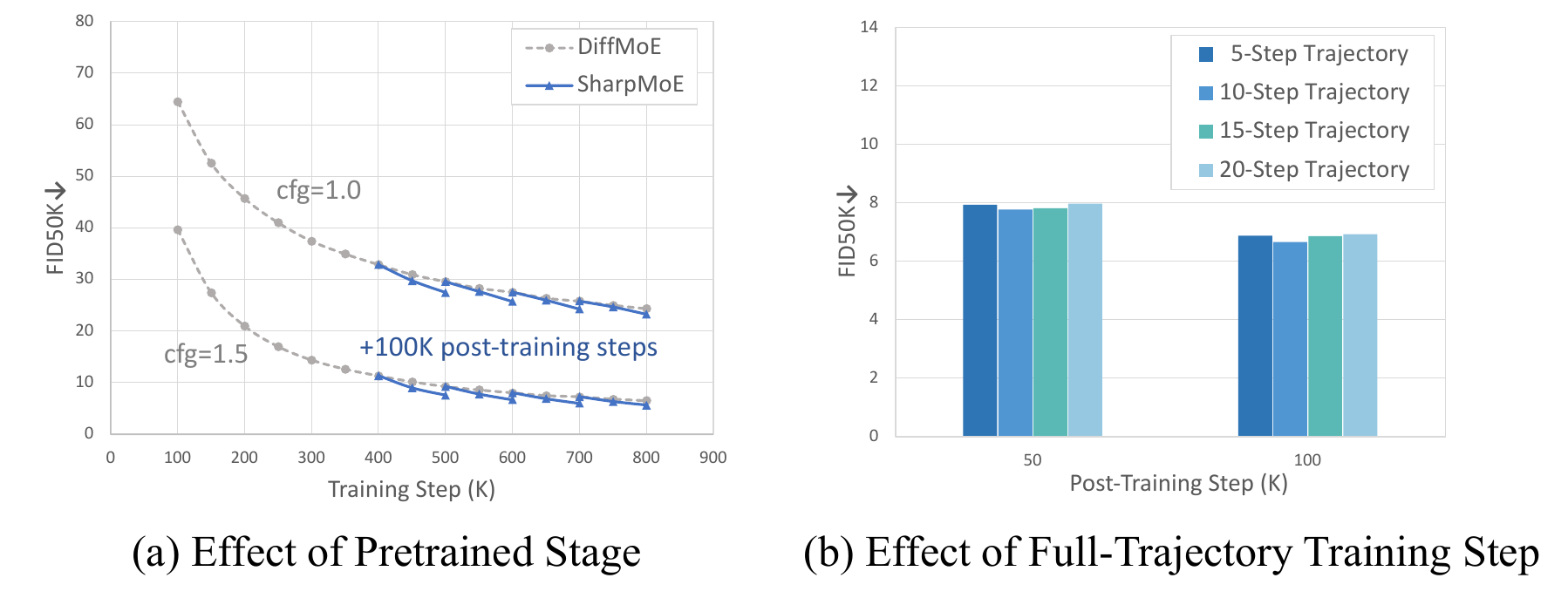}
    \vspace{-1em}
    \caption{
    Ablation studies on our critical designs. (a) Effect of Pretrained Stage. SharpMoE consistently boosts various DiffMoE-B checkpoints within only 100K post-training steps. (b) Effect of Training Trajectory Step $T$. SharpMoE consistently delivers substantial performance gains across various rollout steps, showing high robustness to $T$.
    }
    \vspace{-1em}
    \label{fig:ablation}
\end{figure}

We evaluate the adaptability of SharpMoE by integrating it into DiffMoE-B checkpoints at different training stages. As shown in Fig.~\ref{fig:ablation}(a), SharpMoE consistently achieves a consistent improvement compared to the baseline's original training, regardless of whether it is initialized from a 400K-step or 700K-step pretrained checkpoint. This consistent acceleration in performance underscores SharpMoE's robustness as a plug-and-play enhancement capable of refining expert allocation at any stage of model convergence, even in an already converged state. Remarkably, the fidelity improvements realized within just 100K post-training steps highlight the superior effectiveness of our saliency-harnessing routing as a post-training enhancement.

% surpass those obtained through the baseline's extensive dense training

\paragraph{Effect of Full-Trajectory Training Step.}
We investigate the impact of the rollout step count $T$ within our recursive full-trajectory training scheme.
As illustrated in Fig.~\ref{fig:ablation}(b), SharpMoE exhibits remarkable robustness to the choice of $T$, consistently delivering substantial performance gains over the baseline across a wide range of rollout counts (from $T=5$ to $20$). The marginal performance variations across these settings suggest that the saliency-harnessing routing mechanism provides stable and reliable guidance regardless of the specific training step count. This robustness underscores the practicality and adaptability of SharpMoE, as it can be effectively deployed without the need for exhaustive hyperparameter tuning. Among these effective configurations, we empirically select $T=10$ for our experiments, as it delivers the highest generative fidelity across the spectrum.

\begin{figure}[t]
    \centering
    \includegraphics[width=1.00\linewidth]{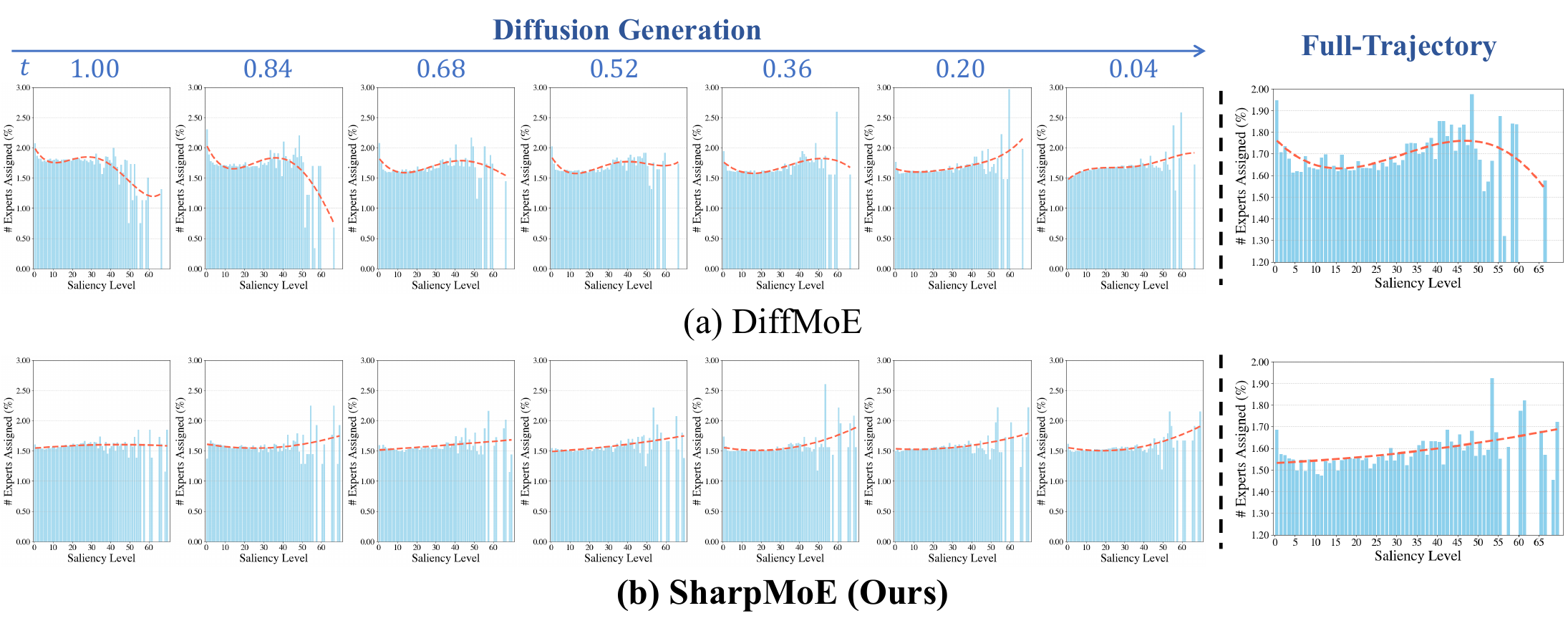}
    \vspace{-2em}
    \caption{
    Aggregated distribution of saliency levels and the number of assigned experts averaged across multiple generated images: (Left) Per-timestep during generation and (Right) Full trajectory. (a) DiffMoE exhibits saliency-insensitive allocation due to the limitations of noisy routing. (b) SharpMoE forms a strong monotonic correlation, prioritizing salient regions, with more notable gains in high-noise stages.
    }
    \vspace{-0.5em}
    \label{fig:result_expert_sal}
\end{figure}

\paragraph{Visualization of Expert Allocation.}
To gain deeper insights into the routing behavior, we visualize the relationship between token saliency and expert assignment of each timestep and the full trajectory in Fig.~\ref{fig:result_expert_sal}.
Using the Laplacian response of the target image as the saliency level, we track the average number of experts assigned to tokens across varying saliency levels within several generated images. Fig.~\ref{fig:result_expert_sal}(a) illustrates that the baseline DiffMoE exhibits saliency-insensitive allocation, with expert assignments largely uncorrelated with the textural complexity of the tokens, especially at early stages where noise levels are high. This empirically confirms the previously identified routing assignment issue: routers conditioned on noise-corrupted latents struggle to distinguish salient regions from background areas.

In contrast, as shown in Fig.~\ref{fig:result_expert_sal}(b), SharpMoE establishes a clear monotonic relationship between token saliency and allocated computational resources. The upward-sloping trend indicates that tokens with higher structural or textural richness are consistently assigned more experts. Importantly, the improvement in routing accuracy is most pronounced during high-noise stages, where SharpMoE successfully harnesses the noise-free latent saliency to guide resource allocation. This alignment underscores the effectiveness of the proposed saliency-harnessing mechanism, which leverages clean latent guidance to provide the router with a stable, noise-free saliency signal. By precisely prioritizing salient tokens, SharpMoE efficiently allocates computational resources to regions critical for generative fidelity, thereby supporting the observed improvements in visual generation.

\vspace{-0.25em}
\section{Conclusion}
\vspace{-0.25em}

We present SharpMoE, a post-training framework for diffusion MoE that tackles the routing assignment problem: existing routers conditioned on noisy latents struggle to recognize salient tokens, leading to saliency-insensitive compute allocation.
Specifically, SharpMoE introduces a saliency-harnessing accurate routing mechanism that employs the clean prediction as the saliency representation, thereby facilitating noise-free routing.
Building upon the full-trajectory training scheme, we further propose a trajectory routing loss that aligns the cumulative expert assignment along the denoising rollout with the saliency distribution, enabling saliency-aware resource prioritization.
Extensive experiments across multiple diffusion MoEs demonstrate that SharpMoE is plug-and-play, requires only lightweight post-training, and consistently enhances generation quality.

\section*{Acknowledgements}

This work is supported by the National Natural Science Foundation of China under grant U22B2053.

\bibliographystyle{splncs04}
\bibliography{main}

@String(CVPR  = {IEEE Conf. Comput. Vis. Pattern Recog.})

@String(ICCV  = {Int. Conf. Comput. Vis.})

@String(ECCV  = {Eur. Conf. Comput. Vis.})

@String(NeurIPS = {Adv. Neural Inform. Process. Syst.})

@String(ICML  = {Int. Conf. Mach. Learn.})

@String(MICCAI  = {Int. Conf. Med. Image Comput. Comput.-Assist. Interv.})

@String(CVPR  = {CVPR})

@String(ICCV  = {ICCV})

@String(ECCV  = {ECCV})

@String(NeurIPS = {NeurIPS})

@String(ICML  = {ICML})

@String(MICCAI  = {MICCAI})

@inproceedings{ronneberger2015unet,
  title={U-net: Convolutional networks for biomedical image segmentation},
  author={Ronneberger, Olaf and Fischer, Philipp and Brox, Thomas},
  booktitle=MICCAI,
  pages={234--241},
  year={2015},
  organization={Springer}
}

@article{ho2020denoising,
  title={Denoising diffusion probabilistic models},
  author={Ho, Jonathan and Jain, Ajay and Abbeel, Pieter},
  journal=NeurIPS,
  volume={33},
  pages={6840--6851},
  year={2020}
}

@article{podell2023sdxl,
  title={Sdxl: Improving latent diffusion models for high-resolution image synthesis},
  author={Podell, Dustin and English, Zion and Lacey, Kyle and Blattmann, Andreas and Dockhorn, Tim and M{\"u}ller, Jonas and Penna, Joe and Rombach, Robin},
  journal={arXiv preprint arXiv:2307.01952},
  year={2023}
}

@article{song2020score,
  title={Score-based generative modeling through stochastic differential equations},
  author={Song, Yang and Sohl-Dickstein, Jascha and Kingma, Diederik P and Kumar, Abhishek and Ermon, Stefano and Poole, Ben},
  journal={arXiv preprint arXiv:2011.13456},
  year={2020}
}

@inproceedings{rombach2022high,
  title={High-resolution image synthesis with latent diffusion models},
  author={Rombach, Robin and Blattmann, Andreas and Lorenz, Dominik and Esser, Patrick and Ommer, Bj{\"o}rn},
  booktitle=CVPR,
  pages={10684--10695},
  year={2022}
}

@inproceedings{peebles2023scalable,
  title={Scalable diffusion models with transformers},
  author={Peebles, William and Xie, Saining},
  booktitle=ICCV,
  pages={4195--4205},
  year={2023}
}

@article{deng2026densegrpo,
  title={Densegrpo: From sparse to dense reward for flow matching model alignment},
  author={Deng, Haoyou and Yan, Keyu and Mao, Chaojie and Wang, Xiang and Liu, Yu and Gao, Changxin and Sang, Nong},
  journal={arXiv preprint arXiv:2601.20218},
  year={2026}
}

@article{wang2025hbridge,
  title={HBridge: H-Shape Bridging of Heterogeneous Experts for Unified Multimodal Understanding and Generation},
  author={Wang, Xiang and Zhang, Zhifei and Zhang, He and Lin, Zhe and Zhou, Yuqian and Liu, Qing and Zhang, Shiwei and Li, Yijun and Liu, Shaoteng and Zheng, Haitian and others},
  journal={arXiv preprint arXiv:2511.20520},
  year={2025}
}

@article{sun2024ecdit,
  title={Ec-dit: Scaling diffusion transformers with adaptive expert-choice routing},
  author={Sun, Haotian and Lei, Tao and Zhang, Bowen and Li, Yanghao and Huang, Haoshuo and Pang, Ruoming and Dai, Bo and Du, Nan},
  journal={arXiv preprint arXiv:2410.02098},
  year={2024}
}

@article{fei2024tcdit,
  title={Scaling diffusion transformers to 16 billion parameters},
  author={Fei, Zhengcong and Fan, Mingyuan and Yu, Changqian and Li, Debang and Huang, Junshi},
  journal={arXiv preprint arXiv:2407.11633},
  year={2024}
}

@article{shi2025diffmoe,
  title={Diffmoe: Dynamic token selection for scalable diffusion transformers},
  author={Shi, Minglei and Yuan, Ziyang and Yang, Haotian and Wang, Xintao and Zheng, Mingwu and Tao, Xin and Zhao, Wenliang and Zheng, Wenzhao and Zhou, Jie and Lu, Jiwen and others},
  journal={arXiv preprint arXiv:2503.14487},
  year={2025}
}

@article{yuan2025expertrace,
  title={Expert race: A flexible routing strategy for scaling diffusion transformer with mixture of experts},
  author={Yuan, Yike and Wang, Ziyu and Huang, Zihao and Zhu, Defa and Zhou, Xun and Yu, Jingyi and Min, Qiyang},
  journal={arXiv preprint arXiv:2503.16057},
  year={2025}
}

@article{liu2022rectifiedflow,
  title={Flow straight and fast: Learning to generate and transfer data with rectified flow},
  author={Liu, Xingchao and Gong, Chengyue and Liu, Qiang},
  journal={arXiv preprint arXiv:2209.03003},
  year={2022}
}

@inproceedings{wei2024dreamvideo,
  title={Dreamvideo: Composing your dream videos with customized subject and motion},
  author={Wei, Yujie and Zhang, Shiwei and Qing, Zhiwu and Yuan, Hangjie and Liu, Zhiheng and Liu, Yu and Zhang, Yingya and Zhou, Jingren and Shan, Hongming},
  booktitle=CVPR,
  pages={6537--6549},
  year={2024}
}

@inproceedings{hatamizadeh2024diffit,
  title={Diffit: Diffusion vision transformers for image generation},
  author={Hatamizadeh, Ali and Song, Jiaming and Liu, Guilin and Kautz, Jan and Vahdat, Arash},
  booktitle=ECCV,
  pages={37--55},
  year={2024},
  organization={Springer}
}

@article{chen2023pixart,
  title={Pixart-$\alpha$: Fast training of diffusion transformer for photorealistic text-to-image synthesis},
  author={Chen, Junsong and Yu, Jincheng and Ge, Chongjian and Yao, Lewei and Xie, Enze and Wu, Yue and Wang, Zhongdao and Kwok, James and Luo, Ping and Lu, Huchuan and others},
  journal={arXiv preprint arXiv:2310.00426},
  year={2023}
}

@inproceedings{ma2024sit,
  title={Sit: Exploring flow and diffusion-based generative models with scalable interpolant transformers},
  author={Ma, Nanye and Goldstein, Mark and Albergo, Michael S and Boffi, Nicholas M and Vanden-Eijnden, Eric and Xie, Saining},
  booktitle=ECCV,
  pages={23--40},
  year={2024},
  organization={Springer}
}

@article{jacobs1991adaptive,
  title={Adaptive mixtures of local experts},
  author={Jacobs, Robert A and Jordan, Michael I and Nowlan, Steven J and Hinton, Geoffrey E},
  journal={Neural computation},
  volume={3},
  number={1},
  pages={79--87},
  year={1991},
  publisher={MIT Press}
}

@article{shazeer2017outrageously,
  title={Outrageously large neural networks: The sparsely-gated mixture-of-experts layer},
  author={Shazeer, Noam and Mirhoseini, Azalia and Maziarz, Krzysztof and Davis, Andy and Le, Quoc and Hinton, Geoffrey and Dean, Jeff},
  journal={arXiv preprint arXiv:1701.06538},
  year={2017}
}

@article{wei2025routing,
  title={Routing matters in moe: Scaling diffusion transformers with explicit routing guidance},
  author={Wei, Yujie and Zhang, Shiwei and Yuan, Hangjie and Han, Yujin and Chen, Zhekai and Wang, Jiayu and Zou, Difan and Liu, Xihui and Zhang, Yingya and Liu, Yu and others},
  journal={arXiv preprint arXiv:2510.24711},
  year={2025}
}

@article{liu2024deepseek,
  title={Deepseek-v3 technical report},
  author={Liu, Aixin and Feng, Bei and Xue, Bing and Wang, Bingxuan and Wu, Bochao and Lu, Chengda and Zhao, Chenggang and Deng, Chengqi and Zhang, Chenyu and Ruan, Chong and others},
  journal={arXiv preprint arXiv:2412.19437},
  year={2024}
}

@article{li2025minimax,
  title={Minimax-01: Scaling foundation models with lightning attention},
  author={Li, Aonian and Gong, Bangwei and Yang, Bo and Shan, Boji and Liu, Chang and Zhu, Cheng and Zhang, Chunhao and Guo, Congchao and Chen, Da and Li, Dong and others},
  journal={arXiv preprint arXiv:2501.08313},
  year={2025}
}

@article{lepikhin2020gshard,
  title={Gshard: Scaling giant models with conditional computation and automatic sharding},
  author={Lepikhin, Dmitry and Lee, HyoukJoong and Xu, Yuanzhong and Chen, Dehao and Firat, Orhan and Huang, Yanping and Krikun, Maxim and Shazeer, Noam and Chen, Zhifeng},
  journal={arXiv preprint arXiv:2006.16668},
  year={2020}
}

@article{lipman2022flow,
  title={Flow matching for generative modeling},
  author={Lipman, Yaron and Chen, Ricky TQ and Ben-Hamu, Heli and Nickel, Maximilian and Le, Matt},
  journal={arXiv preprint arXiv:2210.02747},
  year={2022}
}

@inproceedings{deng2009imagenet,
  title={Imagenet: A large-scale hierarchical image database},
  author={Deng, Jia and Dong, Wei and Socher, Richard and Li, Li-Jia and Li, Kai and Fei-Fei, Li},
  booktitle=CVPR,
  pages={248--255},
  year={2009},
  organization={Ieee}
}

@article{heusel2017gans,
  title={Gans trained by a two time-scale update rule converge to a local nash equilibrium},
  author={Heusel, Martin and Ramsauer, Hubert and Unterthiner, Thomas and Nessler, Bernhard and Hochreiter, Sepp},
  journal=NeurIPS,
  volume={30},
  year={2017}
}

@article{dhariwal2021diffusion,
  title={Diffusion models beat gans on image synthesis},
  author={Dhariwal, Prafulla and Nichol, Alexander},
  journal=NeurIPS,
  volume={34},
  pages={8780--8794},
  year={2021}
}

@article{salimans2016improved,
  title={Improved techniques for training gans},
  author={Salimans, Tim and Goodfellow, Ian and Zaremba, Wojciech and Cheung, Vicki and Radford, Alec and Chen, Xi},
  journal=NeurIPS,
  volume={29},
  year={2016}
}

@article{dai2024deepseekmoe,
  title={DeepSeekMoE: Towards Ultimate Expert Specialization in Mixture-of-Experts Language Models},
  author={Dai, Damai and Deng, Chengqi and Zhao, Chenggang and Xu, RX and Gao, Huazuo and Chen, Deli and Li, Jiashi and Zeng, Wangding and Yu, Xingkai and Wu, Y and others},
  journal={arXiv preprint arXiv:2401.06066},
  year={2024}
}

@article{wang2024auxiliary,
  title={Auxiliary-loss-free load balancing strategy for mixture-of-experts},
  author={Wang, Lean and Gao, Huazuo and Zhao, Chenggang and Sun, Xu and Dai, Damai},
  journal={arXiv preprint arXiv:2408.15664},
  year={2024}
}

@article{zhou2022mixture,
  title={Mixture-of-experts with expert choice routing},
  author={Zhou, Yanqi and Lei, Tao and Liu, Hanxiao and Du, Nan and Huang, Yanping and Zhao, Vincent and Dai, Andrew M and Le, Quoc V and Laudon, James and others},
  journal=NeurIPS,
  volume={35},
  pages={7103--7114},
  year={2022}
}

@article{xue2023raphael,
  title={Raphael: Text-to-image generation via large mixture of diffusion paths},
  author={Xue, Zeyue and Song, Guanglu and Guo, Qiushan and Liu, Boxiao and Zong, Zhuofan and Liu, Yu and Luo, Ping},
  journal=NeurIPS,
  volume={36},
  pages={41693--41706},
  year={2023}
}

@article{zhao2024dynamic,
  title={Dynamic diffusion transformer},
  author={Zhao, Wangbo and Han, Yizeng and Tang, Jiasheng and Wang, Kai and Song, Yibing and Huang, Gao and Wang, Fan and You, Yang},
  journal={arXiv preprint arXiv:2410.03456},
  year={2024}
}

@article{wang2024remoe,
  title={Remoe: Fully differentiable mixture-of-experts with relu routing},
  author={Wang, Ziteng and Zhu, Jun and Chen, Jianfei},
  journal={arXiv preprint arXiv:2412.14711},
  year={2024}
}

@article{ho2022classifier,
  title={Classifier-free diffusion guidance},
  author={Ho, Jonathan and Salimans, Tim},
  journal={arXiv preprint arXiv:2207.12598},
  year={2022}
}

@article{balaji2022ediff,
  title={ediff-i: Text-to-image diffusion models with an ensemble of expert denoisers},
  author={Balaji, Yogesh and Nah, Seungjun and Huang, Xun and Vahdat, Arash and Song, Jiaming and Zhang, Qinsheng and Kreis, Karsten and Aittala, Miika and Aila, Timo and Laine, Samuli and others},
  journal={arXiv preprint arXiv:2211.01324},
  year={2022}
}

@article{wan2025wan,
  title={Wan: Open and advanced large-scale video generative models},
  author={Wan, Team and Wang, Ang and Ai, Baole and Wen, Bin and Mao, Chaojie and Xie, Chen-Wei and Chen, Di and Yu, Feiwu and Zhao, Haiming and Yang, Jianxiao and others},
  journal={arXiv preprint arXiv:2503.20314},
  year={2025}
}

@article{seedream2025seedream,
  title={Seedream 4.0: Toward next-generation multimodal image generation},
  author={Seedream, Team and Chen, Yunpeng and Gao, Yu and Gong, Lixue and Guo, Meng and Guo, Qiushan and Guo, Zhiyao and Hou, Xiaoxia and Huang, Weilin and Huang, Yixuan and others},
  journal={arXiv preprint arXiv:2509.20427},
  year={2025}
}

@article{wu2025qwen,
  title={Qwen-image technical report},
  author={Wu, Chenfei and Li, Jiahao and Zhou, Jingren and Lin, Junyang and Gao, Kaiyuan and Yan, Kun and Yin, Sheng-ming and Bai, Shuai and Xu, Xiao and Chen, Yilei and others},
  journal={arXiv preprint arXiv:2508.02324},
  year={2025}
}

@inproceedings{esser2024scaling,
  title={Scaling rectified flow transformers for high-resolution image synthesis},
  author={Esser, Patrick and Kulal, Sumith and Blattmann, Andreas and Entezari, Rahim and M{\"u}ller, Jonas and Saini, Harry and Levi, Yam and Lorenz, Dominik and Sauer, Axel and Boesel, Frederic and others},
  booktitle=ICML,
  year={2024}
}

@article{cao2025hunyuanimage,
  title={Hunyuanimage 3.0 technical report},
  author={Cao, Siyu and Chen, Hangting and Chen, Peng and Cheng, Yiji and Cui, Yutao and Deng, Xinchi and Dong, Ying and Gong, Kipper and Gu, Tianpeng and Gu, Xiusen and others},
  journal={arXiv preprint arXiv:2509.23951},
  year={2025}
}

@article{mao2026wanimage,
  title={Wan-image: Pushing the boundaries of generative visual intelligence},
  author={Mao, Chaojie and Xie, Chen-Wei and Zhong, Chongyang and Deng, Haoyou and Zhao, Jiaxing and Xiao, Jie and Xing, Jinbo and Zhang, Jingfeng and Zhou, Jingren and Zhang, Jingyi and others},
  journal={arXiv preprint arXiv:2604.19858},
  year={2026}
}
\end{document}